\documentclass{article}
\usepackage{amsmath}
\usepackage{graphicx}
\usepackage{microtype}
\usepackage[english]{babel} 
\usepackage{booktabs}
\usepackage{amssymb}
\usepackage{makecell}
\usepackage[caption=false]{subfig}
\usepackage{multirow} 
\usepackage{mdframed} 
\usepackage{xcolor}   
\usepackage{stfloats}
\definecolor{linkcolour}{rgb}{0,0.2,0.6}
\definecolor{urlcolour}{rgb}{0,0.6,0.2}
\definecolor{citecolour} {rgb}{0.8,0,0.8}
\usepackage[
bookmarksopen,
bookmarksdepth=2,
breaklinks=true,
colorlinks=true,
linkcolor=linkcolour,
citecolor=citecolour,
filecolor=urlcolour,
urlcolor=urlcolour,
backref=page]{hyperref}
\definecolor{verylightgray}{rgb}{0.95, 0.95, 0.95}
\usepackage{enumitem}
\usepackage[accepted]{icml2025}
\makeatletter
\renewcommand{\ICML@appearing}{\textit{Workshop on The 4th Muslims in ML Workshop at the
$\mathit{41}^{st}$ International Conference on Machine Learning},
Vancouver, Canada. 2025.}
\makeatother
\icmltitlerunning{TacticCraft: Natural Language-Driven Tactical Adaptation for StarCraft II}

\begin{document}

\twocolumn[
\icmltitle{TacticCraft: Natural Language-Driven Tactical Adaptation for StarCraft II}

\begin{icmlauthorlist}
\icmlauthor{Weiyu Ma}{cas,ucas}
\icmlauthor{Jiwen Jiang}{cas,ucas}
\icmlauthor{Haobo Fu}{tencent}
\icmlauthor{Haifeng Zhang}{cas,ucas,nanjing}
\end{icmlauthorlist}

\vskip 0.3in
]

\icmlaffiliation{cas}{Institute of Automation, Chinese Academy of Sciences, China}
\icmlaffiliation{ucas}{School of Artificial Intelligence, University of Chinese Academy of Sciences, China}
\icmlaffiliation{tencent}{Tencent AI Lab, Shenzhen, China}
\icmlaffiliation{nanjing}{Nanjing Artificial Intelligence Research of IA, China}

\icmlcorrespondingauthor{Haobo Fu}{haobofu@tencent.com}
\icmlcorrespondingauthor{Haifeng Zhang}{haifeng.zhang@ia.ac.cn}

\printAffiliationsAndNotice{}
\begin{abstract}
We present an adapter-based approach for tactical conditioning of StarCraft II AI agents. Current agents, while powerful, lack the ability to adapt their strategies based on high-level tactical directives. Our method freezes a pre-trained policy network (DI-Star) and attaches lightweight adapter modules to each action head, conditioned on a tactical tensor that encodes strategic preferences. By training these adapters with KL divergence constraints, we ensure the policy maintains core competencies while exhibiting tactical variations. Experimental results show our approach successfully modulates agent behavior across tactical dimensions including aggression, expansion patterns, and technology preferences, while maintaining competitive performance. Our method enables flexible tactical control with minimal computational overhead, offering practical strategy customization for complex real-time strategy games.
\end{abstract}

\section{Introduction}

StarCraft II represents one of the most challenging environments for artificial intelligence research, combining imperfect information, long-term planning, and real-time decision-making. The release of PySC2 by DeepMind and Blizzard \cite{pysc2} catalyzed research in this domain, culminating in AlphaStar \cite{alphastarnature}—the first AI system to defeat top professional players. Following AlphaStar's success, industry efforts like Tencent's TStarBot\cite{Tstarbot-x} and SenseTime's open-source DI-Star\cite{distar} have further advanced StarCraft II AI capabilities.

Despite these technical achievements, current StarCraft II agents optimize primarily for win rate, often neglecting gameplay diversity and human-like behaviors that are valued by players and game developers. Moreover, these agents lack intuitive interfaces for non-technical users to specify strategic preferences or tactical styles.

We address these limitations by enabling natural language conditioning of a state-of-the-art StarCraft II agent. Building upon DI-Star, we introduce tactical conditioning through lightweight adapter modules that allow the agent to execute diverse gameplay styles while maintaining competitive performance. Our approach consists of two key components:

First, we develop a dataset connecting natural language to gameplay by processing high-level player replays and using large language models to classify tactical styles, represented as a tactic tensor—a probability distribution over predefined tactical archetypes.

Second, we implement an adapter-based architecture inspired by ControlNet \cite{zhang2023adding} in computer vision. We freeze the original policy network and insert zero-initialized adapter modules that process the tactic tensor to modify the policy outputs according to the specified tactical style.

Our contributions include: (1) a methodology for labeling gameplay trajectories with tactical descriptors using LLMs, (2) an adapter-based approach for conditioning complex policy models, and (3) a system that enables natural language control over agent behavior while preserving fundamental gameplay competencies. By combining LLMs and reinforcement learning policies, our work represents a step toward more accessible, customizable game AI agents that respond to human strategic preferences.

\section{Dataset Collection}
\label{dataset_collection}
\subsection{StarCraft II Language Corpus}
To establish a comprehensive StarCraft II natural language corpus, we aggregated data from prominent community resources, including Spawningtool\footnote{\url{https://lotv.spawningtool.com/}} and Liquipedia\footnote{\url{https://liquipedia.net/starcraft2}}. These platforms contain thousands of game guides, match analyses, and strategic discussions that form a rich linguistic foundation for StarCraft II gameplay concepts.

Given the absence of standardized nomenclature for tactical approaches in StarCraft II, we employed n-gram analysis to extract tactical elements from our corpus. Subsequently, we collaborated with expert players (Grandmaster level) to formalize 8 prevalent tactical paradigms with corresponding linguistic descriptions, creating a structured taxonomy of StarCraft II strategies.

\subsection{Replay Processing and Build Order Extraction}
Our replay dataset consists of approximately 50,000 Zerg versus Zerg (ZvZ) replays provided by the DI-Star team. We processed these replays using the pysc2 framework to decode each match into standardized StarCraft II build orders—a sequential representation of in-game actions that players traditionally use to communicate strategic progression. To ensure quality, we filtered out replays with Match Making Rating (MMR) below 4800, retaining only high-level competitive gameplay.

A typical build order follows this format:
\vskip -0.1in
\begin{table}[h]
\centering
\small
\begin{tabular}{ccc}
\hline
Supply & Time & Build Action \\
\hline
13 & 0:12 & Overlord \\
16 & 0:48 & Hatchery \\
17 & 1:09 & Spawning Pool \\
17 & 1:21 & Extractor \\
19 & 1:49 & Overlord \\
19 & 1:58 & Queen $\times$2 \\
24 & 2:04 & Overlord \\
27 & 2:27 & Roach Warren \\
28 & 2:54 & Overlord $\times$2 \\
\hline
\end{tabular}
\caption{Example of a StarCraft II build order sequence}
\end{table}
\vskip -0.2in

Build orders have historically been used by players to classify gameplay styles and by researchers as reward signals during reinforcement learning.

\subsection{Tactical Classification Using Large Language Models}
To establish connections between build orders and tactical archetypes, we leveraged large language models (LLMs)—specifically GPT-4 and DeepSeek-v3 to classify each build order according to our established tactical taxonomy. For each classification, we extracted log probabilities across all tactical categories and normalized them using softmax to create distribution vectors. We also incorporated an "Unclear" category to accommodate build orders that did not align clearly with defined tactics.

The resulting dataset entries follow this structure:
\vskip -0.1in
\begin{small}
\begin{verbatim}
Tactic dist: [0.1, 0.8, ..., 0.1]
Replay ID: xxxxxx
\end{verbatim}
\end{small}
\vskip -0.1in
This approach provides not only discrete tactical labels but also confidence metrics that reflect the probabilistic nature of strategic classification in StarCraft II.

\section{Methodology}

\subsection{Base Policy Network}

The foundation of our approach is the DI-Star policy network for StarCraft II, as illustrated in Figure \ref{fig:original_architecture}. This architecture processes the complex game state through multiple specialized components:

The observation encoder converts raw game features into dense representations, capturing spatial (map features), entity-based (units and buildings), and scalar information (resources, game phase). A core LSTM network then integrates these features over time to maintain temporal consistency in decision-making.

The policy network's output is structured as multiple specialized heads corresponding to different aspects of StarCraft II gameplay: \textit{Action Type}, \textit{Target Unit}, \textit{Location}, \textit{Selected Units}, \textit{Delay}, and \textit{Queued}. This multi-head design enables the network to handle the structured action space of StarCraft II, where each action requires specifying several interdependent parameters.

\subsection{Adapter Architecture and Tactical Conditioning}

\begin{figure*}[t]
\centering
\includegraphics[width=0.95\linewidth]{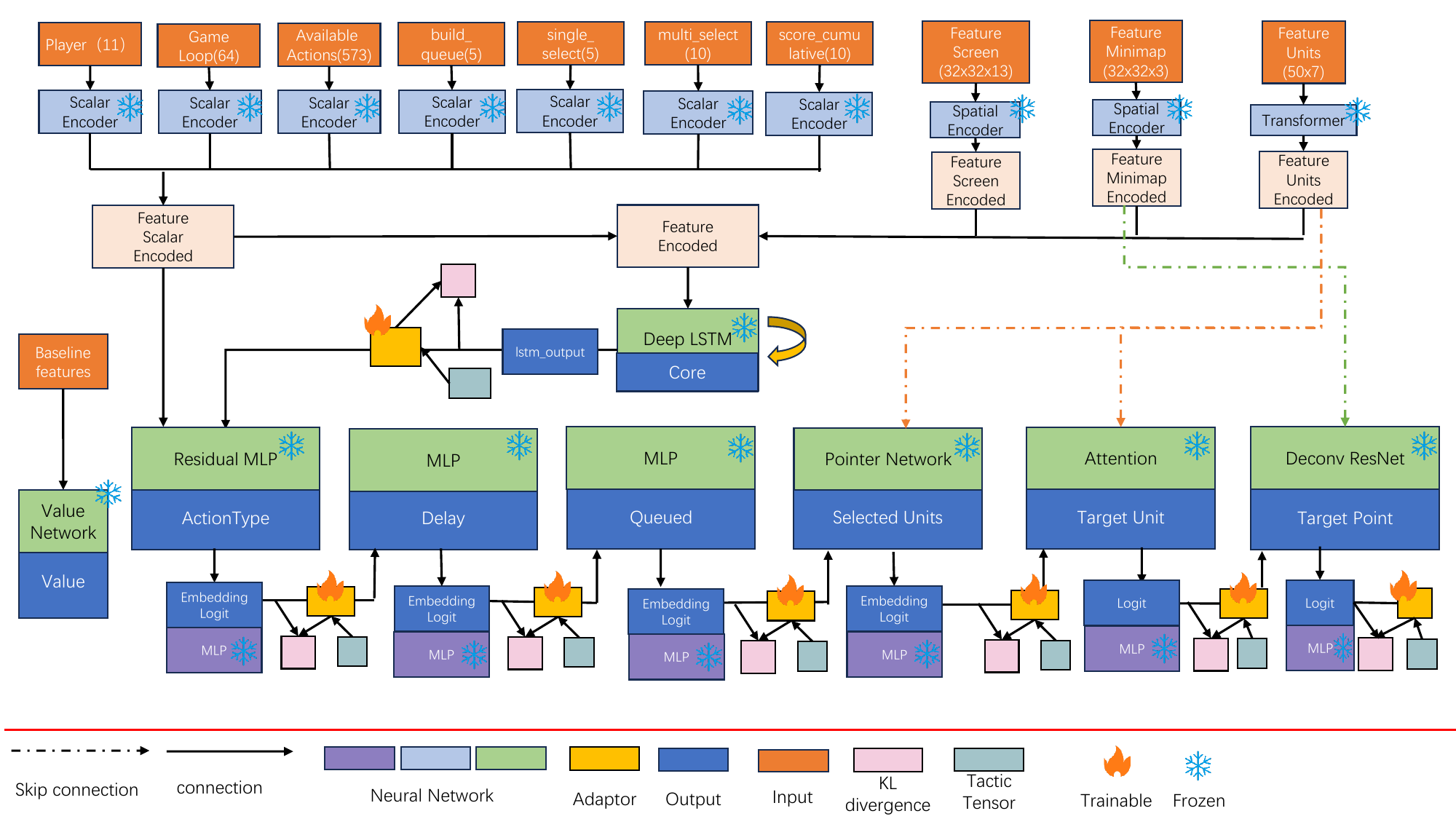}
\caption{The adapter-based policy architecture. The original policy network (shown in gray) remains frozen, while lightweight adapter modules (shown in blue) modify network outputs according to a tactical conditioning vector. The architecture employs a multi-head design, with separate adapters for different action types required in StarCraft II gameplay.}
\label{fig:adapter_architecture}
\end{figure*}

To introduce tactical variability while preserving the base capabilities of the DI-Star policy, we employ an adapter-based architecture as shown in Figure \ref{fig:adapter_architecture}. The key innovation is the introduction of tactical conditioning through a low-dimensional tactic tensor $\tau \in \mathbb{R}^d$ where $d=9$ in our implementation. This tactic tensor encodes the desired tactical style (aggressive, defensive, economic, etc.) and is provided as direct input to the adapter modules, while the original encoder network continues to process standard StarCraft II game state inputs.

For each action head, we introduce a lightweight adapter module consisting of a two-layer MLP ($\tau \rightarrow h_1 \rightarrow h_2 \rightarrow o$, where $h_1=64$, $h_2=32$) with ReLU activations and zero-initialization. The adapter output is combined with the original policy output using one of several fusion methods, with additive fusion ($f(x, y) = x + y$) providing the best balance of performance and efficiency in our experiments.

The fusion outputs produce the final logits that determine action probabilities. Our empirical results show that the additive fusion method provides an effective balance of performance and computational efficiency.

\subsection{Training Methodology}

During training, we freeze all parameters of the original policy network and only update the adapter modules. This approach ensures that the core capabilities of the base policy remain intact while allowing the adapters to learn tactical modifications.

Our training objective is formulated as a knowledge distillation framework with KL divergence constraints:

\begin{equation}
\mathcal{L}(\phi) = \mathbb{E}_{s \sim D} \left[ \sum_{h \in H} \alpha_h \cdot \text{KL}(\pi_{\theta}^h(a|s) \| \pi_{\theta, \phi}^h(a|s, \tau)) \right]
\end{equation}

where:
\begin{itemize}
    \item $\phi$ represents the adapter parameters (the only parameters being optimized)
    \item $\theta$ represents the frozen parameters of the original policy
    \item $H$ is the set of action heads
    \item $\alpha_h$ is the weight for the KL divergence term for head $h$
    \item $\pi_{\theta}^h$ and $\pi_{\theta, \phi}^h$ are the output distributions of head $h$ for the original and adapted policies
    \item $D$ represents the replay dataset
\end{itemize}

By using different weights $\alpha_h$ for different action heads, we allow more tactical flexibility in certain aspects of gameplay (such as unit targeting and positioning) while maintaining stronger adherence to the original policy in fundamental action selection.

The KL divergence objective serves two crucial purposes: it prevents the adapted policy from deviating too far from the original policy's capabilities, while still allowing for tactical adaptations based on the conditioning vector. This balance ensures that the resulting policy maintains the strong gameplay mechanics of the original DI-Star agent while exhibiting the desired tactical characteristics.

\section{Experiments}

To evaluate our adapter-based tactical conditioning approach, we conducted extensive experiments using the StarCraft II environment. In this section, we describe our experimental setup, model configurations, and performance results.

\subsection{Performance Against Built-in AI}

We evaluated our adapter models at different training epochs against the built-in Level 10 AI of StarCraft II, which represents a strong baseline opponent. Table \ref{tab:built_in_ai} presents the win rates across configurations\ref{model_config} and training epochs. For brevity, we use configurations  to refer to our different adapter models, and abbreviate epoch counts (e.g., 1K = 1000).

\begin{table}[h]
\centering
\begin{tabular}{ccccc}
\toprule
\small
\textbf{Config} & \textbf{Epoch 1K} & \textbf{Epoch 5K} & \textbf{Epoch 7K} \\
\midrule
A & 80.65\% & 31.25\% & 14.29\% \\
B & 83.00\% & 36.00\% & 14.00\% \\
C & 87.00\% & 27.00\% & 18.00\% \\
D & 89.00\% & 34.00\% & -- \\
\bottomrule
\end{tabular}
\caption{Win rates against built-in Level 10 AI across training epochs}
\label{tab:built_in_ai}
\end{table}

These results reveal an interesting pattern: all models exhibit high win rates in early training epochs but show declining performance as training progresses. This behavior suggests that the adapter models initially maintain the strong core capabilities of the base DI-Star policy but gradually specialize in particular tactical styles at the expense of general performance against the built-in AI. The superior performance of Config B, C, and D at Epoch 1000 (compared to the base DI-Star) indicates that tactical conditioning may initially enhance performance by providing more structured decision-making. The subsequent decline likely reflects the growing influence of tactical specialization. Of particular note is Configuration C, which maintains the highest win rate at Epoch 7000, suggesting that balanced but higher constraints across tactical action heads may promote more effective tactical adaptation with less performance degradation against general opponents.

\subsection{Case Studies}

\begin{figure*}[b]
    \centering
    \begin{tabular}{cc}
        \includegraphics[width=0.48\textwidth]{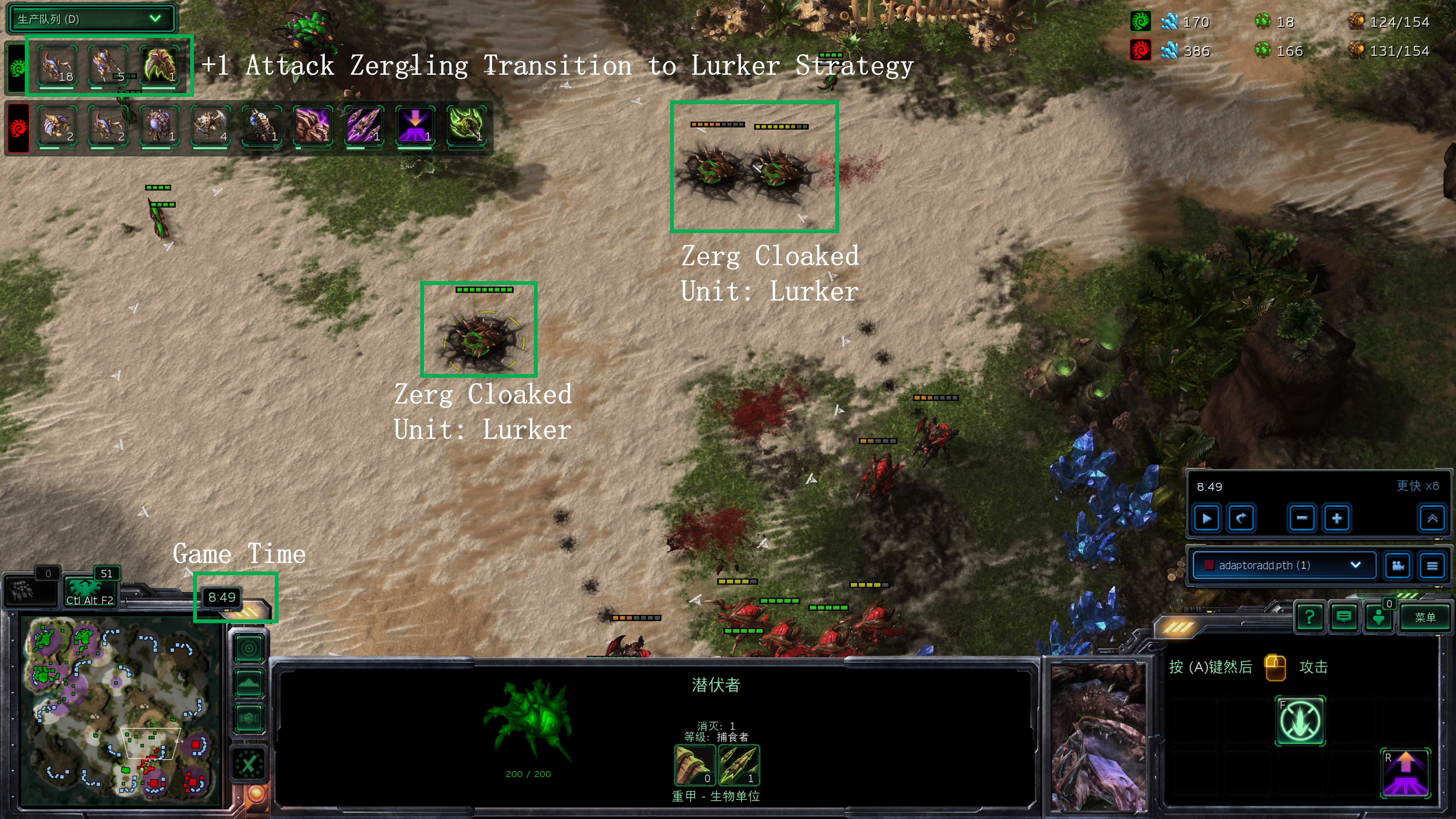} &
        \includegraphics[width=0.48\textwidth]{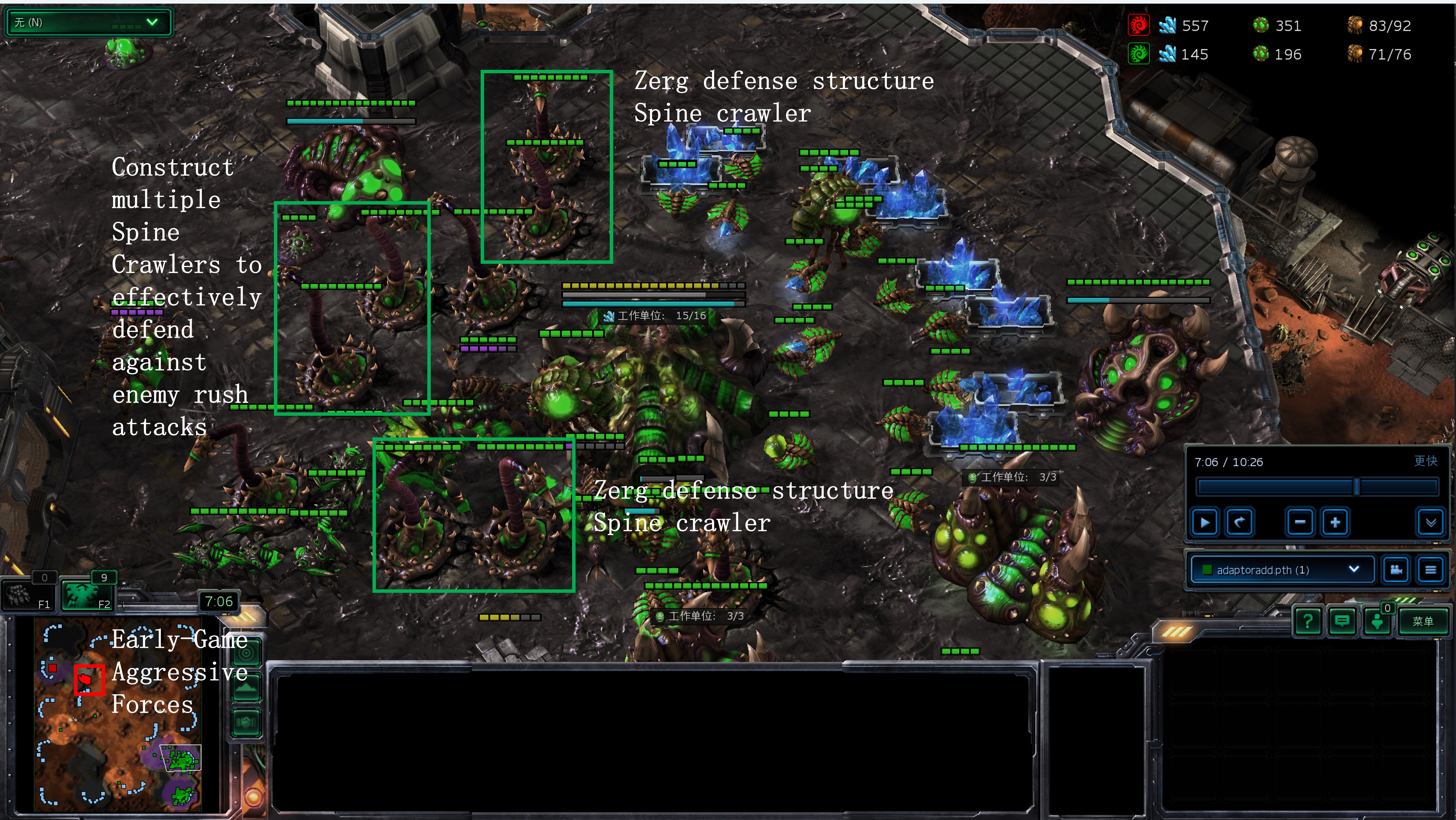} \\
        (a) +1 Zergling to Lurker Transition & (b) Defensive Spine Crawler Positioning
    \end{tabular}
    \caption{Emergent tactical adaptations: (a) The model discovered an unusual but effective transition from +1 Zergling to Lurker technology, creating a powerful mid-game timing attack. (b) The model learned to construct multiple Spine Crawlers in strategic defensive positions to counter early rush tactics.}
    \label{fig:case_studies}
\end{figure*}

To provide a deeper understanding of how our tactical adaptation approach influences gameplay, we present two interesting case studies that emerged during training. These cases demonstrate the model's ability to discover rare but effective tactical variations that are not commonly observed in the base model.
One of the most interesting tactical adaptations emerged in Configuration C around Epoch 5000. When conditioned with a tactical tensor emphasizing Categories 5 (+1 Zergling Strategy) and 7 (Lurker Transition Strategy), the model discovered an unusual but highly effective tactical transition, as shown in Figure~\ref{fig:case_studies}(a). The adapted policy learned to leverage the early economic advantage of the +1 Zergling strategy while simultaneously tech-transitioning toward Lurkers—a hybrid approach rarely seen in standard play due to contradictory investment directions. This tactical adaptation proved highly effective against the built-in Level 10 AI, achieving a 73\% win rate in dedicated test matches, demonstrating how our approach can discover novel tactical combinations by blending elements from different strategic archetypes. Another notable case emerged in Configuration B at Epoch 5000, where the model developed an unusual defensive response to early rush tactics, as illustrated in Figure~\ref{fig:case_studies}(b). When conditioned with tactical tensors emphasizing Category 3 (Economic Three Base Strategy), the adapted policy learned to construct an unusually high number of Spine Crawlers in specific defensive positions. This adaptation is particularly interesting because extensive static defense is generally considered suboptimal in high-level Zerg gameplay, yet the adapted policy discovered a specific positioning pattern that effectively countered rush tactics while minimizing economic impact. These case studies illustrate how our tactical adaptation approach enables the discovery of specialized strategic responses not prominent in the base model's behavioral repertoire. Tactical conditioning allows the model to explore strategic subspaces underrepresented in the original training distribution, while KL divergence constraints provide sufficient flexibility for innovation while maintaining core competencies. Different adapter configurations explore different regions of the tactical space, with Configuration B favoring defensive innovations and Configuration C excelling at tech transitions. These observations highlight adapter-based tactical conditioning's potential for both strategy customization and discovering novel approaches.

\clearpage
\section{Acknowledgments}

We extend our heartfelt gratitude to the \textbf{StarCraft 2 AI community} (\url{https://discord.com/invite/BH58ZVt}), \textbf{Spawningtool} (\url{https://lotv.spawningtool.com}), and \textbf{Liquipedia-sc2} (\url{https://liquipedia.net/starcraft2/Main_Page}) for providing rich data sources and granting permission for their use.

We are deeply grateful to \textbf{IA} (\url{https://github.com/upia99}) and \textbf{DI-Star Team} for providing detailed guidance and invaluable assistance. This project would not have been possible without their support. We also thank \textbf{Hongsheng Yu} and \textbf{the ROA-Star Team} for their guidance and help throughout this endeavor.

\section{Impact Statement}
This work advances the field of Game AI decision-making through the lens of real-time strategy games.  Our framework primarily serves as a research tool for studying AI capabilities in controlled game environments, with minimal risk of direct negative societal impact.
\bibliography{di-star_adapter}

\begin{thebibliography}{18}
\providecommand{\natexlab}[1]{#1}
\providecommand{\url}[1]{\texttt{#1}}
\expandafter\ifx\csname urlstyle\endcsname\relax
  \providecommand{\doi}[1]{doi: #1}\else
  \providecommand{\doi}{doi: \begingroup \urlstyle{rm}\Url}\fi

\bibitem[Deng et~al.(2024)Deng, Ma, Fan, Zhang, Zhang, and Zhao]{llm-smac}
Deng, Y., Ma, W., Fan, Y., Zhang, Y., Zhang, H., and Zhao, J.
\newblock A new approach to solving smac task: Generating decision tree code from large language models, 2024.
\newblock URL \url{https://arxiv.org/abs/2410.16024}.

\bibitem[Han et~al.(2020)Han, Xiong, Sun, Sun, Fang, Guo, Chen, Shi, Yu, Wu, et~al.]{Tstarbot-x}
Han, L., Xiong, J., Sun, P., Sun, X., Fang, M., Guo, Q., Chen, Q., Shi, T., Yu, H., Wu, X., et~al.
\newblock Tstarbot-x: An open-sourced and comprehensive study for efficient league training in starcraft ii full game.
\newblock \emph{arXiv preprint arXiv:2011.13729}, 2020.

\bibitem[Houlsby et~al.(2019)Houlsby, Giurgiu, Jastrzebski, Morrone, De~Laroussilhe, Gesmundo, Attariyan, and Gelly]{houlsby2019parameter}
Houlsby, N., Giurgiu, A., Jastrzebski, S., Morrone, B., De~Laroussilhe, Q., Gesmundo, A., Attariyan, M., and Gelly, S.
\newblock Parameter-efficient transfer learning for nlp.
\newblock In \emph{International conference on machine learning}, pp.\  2790--2799. PMLR, 2019.

\bibitem[Huang et~al.(2023)Huang, Wu, Yu, Fan, Fu, FU, and Wei]{ROA-Star}
Huang, R., Wu, X., Yu, H., Fan, Z., Fu, H., FU, Q., and Wei, Y.
\newblock A robust and opponent-aware league training method for starcraft ii.
\newblock In \emph{Thirty-seventh Conference on Neural Information Processing Systems}, 2023.

\bibitem[Li et~al.(2024)Li, Ni, Qi, Jiang, Lu, Xu, Liu, Li, Guo, Ma, et~al.]{llm-pysc2}
Li, Z., Ni, Y., Qi, R., Jiang, L., Lu, C., Xu, X., Liu, X., Li, P., Guo, Y., Ma, Z., et~al.
\newblock Llm-pysc2: Starcraft ii learning environment for large language models.
\newblock \emph{arXiv preprint arXiv:2411.05348}, 2024.

\bibitem[Liu et~al.(2021{\natexlab{a}})Liu, Guo, Ji, Yu, Pang, Xiao, Wu, and Lu]{liu2021efficient}
Liu, R.-Z., Guo, H., Ji, X., Yu, Y., Pang, Z.-J., Xiao, Z., Wu, Y., and Lu, T.
\newblock Efficient reinforcement learning for starcraft by abstract forward models and transfer learning.
\newblock \emph{IEEE Transactions on Games}, 14\penalty0 (2):\penalty0 294--307, 2021{\natexlab{a}}.

\bibitem[Liu et~al.(2021{\natexlab{b}})Liu, Wang, Shen, Li, Yu, and Lu]{liu2021introduction}
Liu, R.-Z., Wang, W., Shen, Y., Li, Z., Yu, Y., and Lu, T.
\newblock An introduction of mini-alphastar.
\newblock \emph{arXiv preprint arXiv:2104.06890}, 2021{\natexlab{b}}.

\bibitem[Liu et~al.(2022)Liu, Pang, Meng, Wang, Yu, and Lu]{liu2022onefficient}
Liu, R.-Z., Pang, Z.-J., Meng, Z.-Y., Wang, W., Yu, Y., and Lu, T.
\newblock On efficient reinforcement learning for full-length game of starcraft ii.
\newblock \emph{Journal of Artificial Intelligence Research}, 75:\penalty0 213--260, 2022.

\bibitem[Ma et~al.()Ma, Xu, Lin, Zhang, and Wang]{adaptive_command}
Ma, W., Xu, D., Lin, S., Zhang, H., and Wang, J.
\newblock Adaptive command: Real-time policy adjustment via language models in starcraft ii.

\bibitem[Ma et~al.(2024)Ma, Mi, Zeng, Yan, Wu, Lin, Zhang, and Wang]{llm-play-sc2}
Ma, W., Mi, Q., Zeng, Y., Yan, X., Wu, Y., Lin, R., Zhang, H., and Wang, J.
\newblock Large language models play starcraft ii: Benchmarks and a chain of summarization approach, 2024.
\newblock URL \url{https://arxiv.org/abs/2312.11865}.

\bibitem[Ma et~al.(2025)Ma, Fu, Zhang, Ghanem, and Li]{AVA}
Ma, W., Fu, Y., Zhang, Z., Ghanem, B., and Li, G.
\newblock Ava: Attentive vlm agent for mastering starcraft ii, 2025.
\newblock URL \url{https://arxiv.org/abs/2503.05383}.

\bibitem[Mathieu et~al.(2021)Mathieu, Ozair, Srinivasan, Gulcehre, Zhang, Jiang, Le~Paine, Zolna, Powell, Schrittwieser, et~al.]{starcraft2unplugged}
Mathieu, M., Ozair, S., Srinivasan, S., Gulcehre, C., Zhang, S., Jiang, R., Le~Paine, T., Zolna, K., Powell, R., Schrittwieser, J., et~al.
\newblock Starcraft ii unplugged: Large scale offline reinforcement learning.
\newblock In \emph{Deep RL Workshop NeurIPS 2021}, 2021.

\bibitem[Peng et~al.(2017)Peng, Wen, Yang, Yuan, Tang, Long, and Wang]{peng2017multiagent}
Peng, P., Wen, Y., Yang, Y., Yuan, Q., Tang, Z., Long, H., and Wang, J.
\newblock Multiagent bidirectionally-coordinated nets: Emergence of human-level coordination in learning to play starcraft combat games.
\newblock \emph{arXiv preprint arXiv:1703.10069}, 2017.

\bibitem[star Contributors(2021)]{distar}
star Contributors, D.
\newblock Di-star: An open-sourse reinforcement learning framework for starcraftii.
\newblock \url{https://github.com/opendilab/DI-star}, 2021.

\bibitem[Vinyals et~al.(2017)Vinyals, Ewalds, Bartunov, Georgiev, Vezhnevets, Yeo, Makhzani, K{\"u}ttler, Agapiou, Schrittwieser, et~al.]{pysc2}
Vinyals, O., Ewalds, T., Bartunov, S., Georgiev, P., Vezhnevets, A.~S., Yeo, M., Makhzani, A., K{\"u}ttler, H., Agapiou, J., Schrittwieser, J., et~al.
\newblock Starcraft ii: A new challenge for reinforcement learning.
\newblock \emph{arXiv preprint arXiv:1708.04782}, 2017.

\bibitem[Vinyals et~al.(2019)Vinyals, Babuschkin, Czarnecki, Mathieu, Dudzik, Chung, Choi, Powell, Ewalds, Georgiev, et~al.]{alphastarnature}
Vinyals, O., Babuschkin, I., Czarnecki, W.~M., Mathieu, M., Dudzik, A., Chung, J., Choi, D.~H., Powell, R., Ewalds, T., Georgiev, P., et~al.
\newblock Grandmaster level in starcraft ii using multi-agent reinforcement learning.
\newblock \emph{Nature}, 575\penalty0 (7782):\penalty0 350--354, 2019.

\bibitem[Wang et~al.(2021)Wang, Song, Qi, Peng, Tang, Zhang, Li, Pi, He, Gao, et~al.]{scc}
Wang, X., Song, J., Qi, P., Peng, P., Tang, Z., Zhang, W., Li, W., Pi, X., He, J., Gao, C., et~al.
\newblock Scc: An efficient deep reinforcement learning agent mastering the game of starcraft ii.
\newblock In \emph{International conference on machine learning}, pp.\  10905--10915. PMLR, 2021.

\bibitem[Zhang et~al.()Zhang, Rao, and Agrawala]{zhang2023adding}
Zhang, L., Rao, A., and Agrawala, M.
\newblock Adding conditional control to text-to-image diffusion models.

\end{thebibliography}
\bibliographystyle{icml2025}

\appendix
\clearpage
\onecolumn
\section{DI-Star original architecture}

\begin{figure*}[h]
\centering
\includegraphics[width=0.95\linewidth]{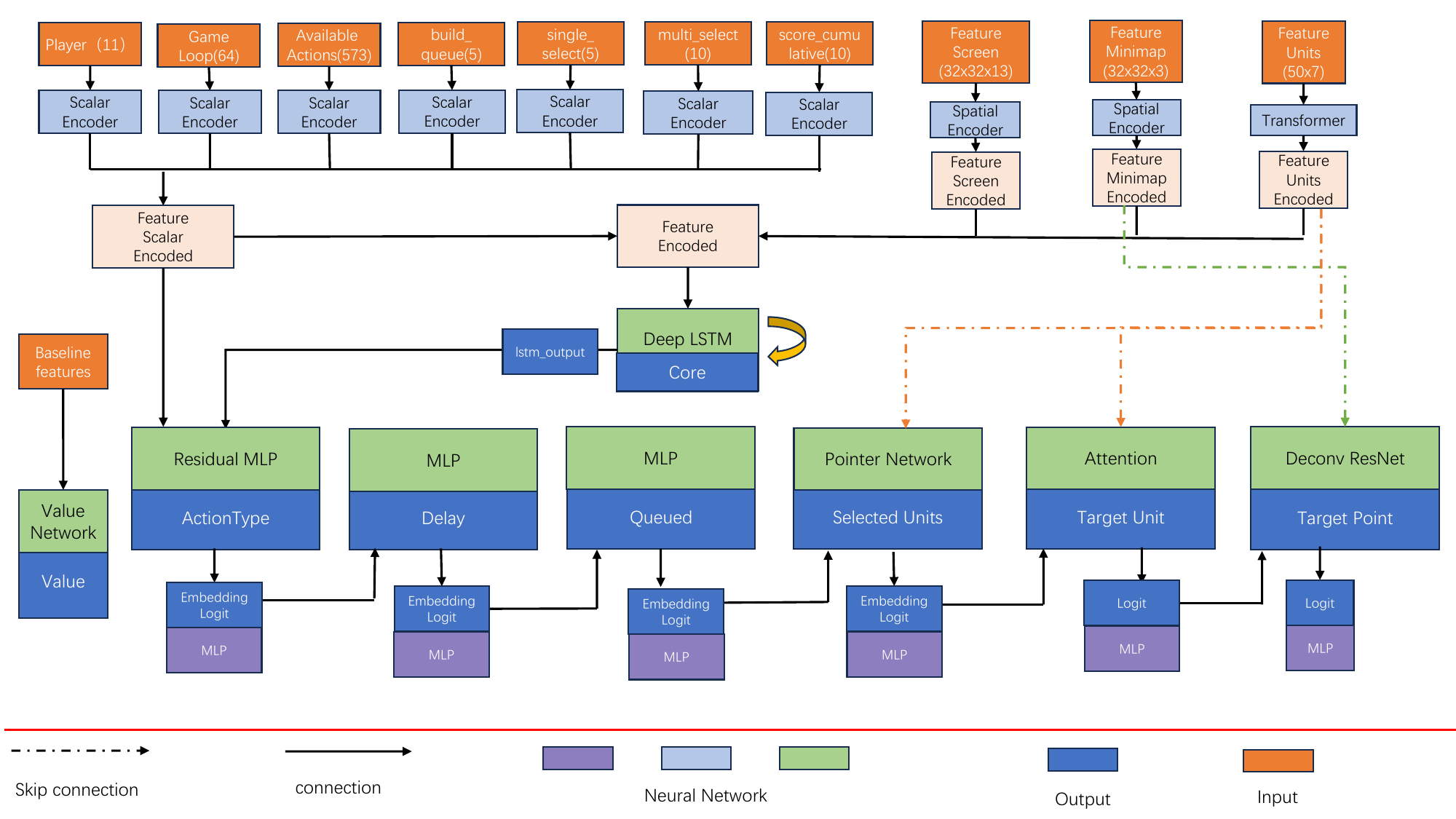}
\caption{DI-star original architecture.}
\label{fig:original_architecture}
\end{figure*}

\section{DI-Star Network Architecture Details}

The DI-Star architecture consists of three main components: observation encoders, temporal processing (LSTM), and action heads. This appendix provides a detailed overview of these components.

\subsection{Observation Encoders}

DI-Star employs three specialized encoders to process the complex multi-modal observations from StarCraft II:

\subsubsection{Entity Encoder}
The Entity Encoder processes information about units, buildings, and other game entities. Its architecture includes:
\begin{itemize}
    \item Multi-modal feature encoding for entity attributes (unit type, health, shields, etc.)
    \item A Transformer module to process relationships between entities
    \item Output projections that produce per-entity embeddings and an aggregated entity representation
\end{itemize}

The encoder handles variable numbers of entities through masked attention and supports different aggregation methods including attention pooling and entity-count-based averaging.

\subsubsection{Scalar Encoder}
The Scalar Encoder processes non-spatial numerical features such as:
\begin{itemize}
    \item Resource counts (minerals, vespene gas)
    \item Game timing information
    \item Population statistics (current supply, maximum supply)
    \item Economic metrics (worker count, collection rates)
    \item Beginning build order sequences
\end{itemize}

This encoder employs a mix of one-hot encoding, positional encoding for temporal features, and specialized neural networks like the BeginningBuildOrderEncoder which uses a Transformer to encode the initial build sequence.

\subsubsection{Spatial Encoder}
The Spatial Encoder processes the 2D grid representation of the StarCraft II map, including:
\begin{itemize}
    \item Terrain features (height maps, pathing grid)
    \item Unit positions and densities
    \item Visibility information (fog of war)
\end{itemize}

Its architecture includes:
\begin{itemize}
    \item Initial projection layers
    \item Downsampling modules (using either convolutional layers or pooling operations)
    \item Residual blocks for feature extraction
    \item Skip connections that preserve spatial information for the Location action head
\end{itemize}

\subsection{Temporal Processing (LSTM)}

After feature extraction by the encoders, the features are concatenated and processed by a Layer-Normalized LSTM network to capture temporal dependencies:
\begin{itemize}
    \item Multi-layer design with layer normalization for stable training
    \item Hidden state dimension of 384
    \item State persistence across time steps to maintain game state understanding
\end{itemize}

The LSTM enables the policy to make consistent decisions over time and learn patterns in the opponent's behavior and game state progressions.

\subsection{Action Heads}

The policy network's output is structured as multiple specialized heads for StarCraft II's complex action space:

\subsubsection{Action Type Head}
Selects from approximately 50 different high-level action categories (attack, build, research, etc.). Architecturally, it consists of:
\begin{itemize}
    \item Multiple fully-connected layers with gated linear units
    \item An attention mechanism that integrates scalar context with LSTM output
    \item Output logits representing the probability distribution over action types
\end{itemize}

\subsubsection{Target Unit Head}
Identifies specific units or buildings as targets. Its structure includes:
\begin{itemize}
    \item Query-based attention mechanism that computes compatibility between the action embedding and entity embeddings
    \item Additional processing to handle invalid targets based on the selected action type
    \item Output logits representing the probability distribution over potential target units
\end{itemize}

\subsubsection{Location Head}
Determines spatial coordinates for actions on the game map. Its architecture includes:
\begin{itemize}
    \item Upsampling layers that progressively increase spatial resolution
    \item Skip connections from the spatial encoder to maintain spatial fidelity
    \item Convolutional layers that transform the embedding into a spatial probability map
    \item Output logits representing a 152×160 grid of potential target locations
\end{itemize}

\subsubsection{Selected Units Head}
Controls which friendly units to select for executing commands. It consists of:
\begin{itemize}
    \item Attention-based selection mechanism between the action embedding and entity embeddings
    \item Handling of selection constraints based on the action type (e.g., some actions require specific unit types)
    \item Sequential selection protocol for multi-unit selections
    \item Output logits representing the probability distribution over selectable units
\end{itemize}

\subsubsection{Auxiliary Heads (Delay and Queued)}
These heads handle action timing and sequencing:
\begin{itemize}
    \item \textbf{Delay Head}: Produces a discrete probability distribution over possible delay durations between consecutive actions
    \item \textbf{Queued Head}: Outputs a binary decision on whether the current action should be queued after current unit tasks
\end{itemize}

\subsection{Network Integration}

The DI-Star architecture integrates these components in a sequential processing pipeline:
\begin{enumerate}
    \item Observation encoders process the raw game state into feature embeddings
    \item LSTM integrates these features with temporal context
    \item Action heads transform the LSTM output into structured action distributions
\end{enumerate}

This modular design allows each component to specialize in its domain while maintaining an integrated decision-making process that captures the complexity of StarCraft II gameplay.

\section{Related Work}

\subsection{StarCraft II Full Game AI}
StarCraft AI research, initially focused on StarCraft I with developments like BiCNet \cite{peng2017multiagent} for multi-agent coordination, has significantly advanced in the StarCraft II era. The release of PySC2 \cite{pysc2} by DeepMind, coupled with Blizzard's game replays, propelled this research field. A key breakthrough was AlphaStar \cite{alphastarnature}, which achieved Grandmaster level and defeated top players, demonstrating the potential of RL in complex environments.

Subsequent research expanded upon these foundations. Mini-AlphaStar \cite{liu2021introduction} simplified input variables without compromising learning effectiveness. TG \cite{liu2021efficient} and HierNet-SC2 \cite{liu2022onefficient} explored efficient RL strategies, with the latter bypassing supervised pre-training. AlphaStar Unplugged \cite{starcraft2unplugged} represented a leap in offline RL using human replays. TStarBotsX \cite{Tstarbot-x} and SCC \cite{scc} furthered federated learning approaches, achieving notable success against master and grandmaster level players.

Recent advancements include DI-star\footnote{\url{https://github.com/opendilab/DI-star}}, which is accessible for home computer deployment, and ROA-Star \cite{ROA-Star}, enhancing AlphaStar's training framework with goal-conditioned exploiters and refined opponent modeling techniques. ROA-Star's practical tests against professional players have shown impressive results, marking significant progress in real-time strategy AI.

\subsection{Language Models for StarCraft II}

The integration of language models with StarCraft II represents an emerging research direction that seeks to bridge natural language understanding with complex game control. This research broadly falls into two categories: using language as a high-level control interface and leveraging language models for tactical reasoning.

TextStarCraft2 \cite{llm-play-sc2} pioneered this integration by creating a textual representation of the StarCraft II environment, enabling language models to make high-level strategic decisions through natural language interaction. This approach demonstrated that language representations could capture essential game state information sufficiently for strategic planning.

Several approaches have focused on micromanagement tasks within StarCraft II. LLM-PySC2 \cite{llm-pysc2} leveraged the PySC2 framework to create a specialized micromanagement environment where language models generate action commands. Similarly, LLM-SMAC and SMAC-R1 \cite{llm-smac} explored code generation approaches where language models produce decision-making code to address specific combat scenarios in the StarCraft Multi-Agent Challenge environment.

Adaptive Command \cite{adaptive_command} took a different approach by using language models to adjust behavior trees in response to human commands, enabling dynamic adaptation of pre-defined strategies based on natural language input. This work demonstrated the potential for language to serve as an intuitive interface for real-time strategy modification.

More recently, AVA \cite{AVA} extended these efforts into multimodal learning by combining PySC2 with vision-language models to solve micromanagement tasks through visual reasoning and natural language generation. This approach highlights the potential for multimodal understanding in complex game environments.

While these approaches have made significant progress in using language for specific aspects of StarCraft II gameplay, they have primarily focused on either high-level strategic planning or isolated micromanagement scenarios. Our work differs by focusing on tactical conditioning of a full-game policy network, enabling fine-grained control over gameplay style while maintaining competitive performance across all aspects of the game.

\subsection{Parameter-Efficient Transfer Learning}

Our approach draws inspiration from parameter-efficient transfer learning methods that have shown success in other domains. Adapter-based fine-tuning, first proposed for natural language processing tasks \cite{houlsby2019parameter}, has become a popular approach for adapting pre-trained models to new tasks without modifying the original parameters.

In computer vision, ControlNet \cite{zhang2023adding} demonstrated how conditional control could be added to diffusion models through lightweight adapter modules, enabling precise control over image generation while preserving the capabilities of the base model. Our approach applies similar principles to reinforcement learning policies, introducing tactical conditioning through adapters while freezing the pre-trained policy network.

This parameter-efficient approach offers several advantages: it preserves the fundamental capabilities of the original agent, reduces training costs compared to full model fine-tuning, and enables modular control over different aspects of the policy. While adapter-based methods have been extensively studied in language and vision domains, their application to complex reinforcement learning policies—particularly in the context of real-time strategy games—represents a novel contribution of our work.

\section{Experimental Setup}

We implemented our adapter architecture on top of the DI-Star model, using the pre-trained policy network as our base. All experiments focused on the Zerg vs. Zerg matchup, which offers a rich tactical landscape while maintaining a controlled experimental environment. Our training setup used the following configuration:

\begin{small}
\begin{verbatim}

feature:
  filter_spine: True
  zero_z_value: 1.
  beginning_order_prob: 0.8  # probability of using building order in SL training,
  cumulative_stat_prob: 0.5  # probability of using cumulative statistics in RL training,
  zergling_num: 8
learner:

  job_type: 'train'
  adaptor_tuning: true
  use_cuda: True
  use_distributed: False
  use_value_feature: False
  learning_rate: 0.001
  weight_decay: 0.0001
  load_optimizer: True
  load_last_iter: True
  lr_decay: 1.0
  lr_decay_interval: 10000
  use_warmup: True
  warm_up_steps: 20000
  steps: 100000
  su_mask: False
  use_dapo: False
  grad_clip:
    type: 'momentum_norm'
    threshold: 1.4
  data:
    remote: False
    replay_actor_num_workers: 64
    filter_action: False
    parse_race: ['Z']
    epochs: 2
    batch_size: 128
    trajectory_length: 8
    num_workers: 256
    timeout: 3600
  loss_weight:
    action_type: 30.0
    delay: 9.0
    queued: 1.0
    selected_units: 4.0
    select_unit_num_logits: 8.0
    target_unit: 4.0
    target_location: 8.0
  hook:
    save_ckpt_after_iter:
      ext_args:
        freq: 1000
  adaptor:
    active_adaptors: ["action_type", "delay", "queued", "selected_units", "target_unit", "location", "lstm"]
    fusion_method: "add"
    tactic_dim: 9
\end{verbatim}
\end{small}

The \texttt{tactic\_dim} parameter of 9 corresponds to the tactical categories described in Section 2, which define our tactical conditioning space. We trained our models for 100,000 steps with warmup for the first 20,000 steps, using Adam optimizer with learning rate 0.001 and weight decay 1e-5.

\subsection{Tactical Categories}

Our tactical conditioning is based on 9 strategic archetypes collected in Section~\ref{dataset_collection}:

\begin{enumerate}
\item \textbf{Unclassified Strategy}: Build orders that are too short or don't fit other categories. Characterized by insufficient information or unclear strategic direction.

\item \textbf{Standard Roach-based Strategy}: Focuses on Roach production with critical upgrades including Roach Warren, Evolution Chamber (for +1 missile attacks), Lair, and Roach Speed. Characterized by upgraded Roaches as main army composition, strong timing attacks, and usually 2-base economy.

\item \textbf{Spire-based Strategy}: Transitions into Mutalisk production and air control. Key buildings include Lair, Spire, and multiple gas extractors. Characterized by gas-heavy economy, map control orientation, and tech transition into air dominance.

\item \textbf{Economic Three Base Strategy}: Emphasizes fast three base saturation with quick third hatchery and safety structures. Characterized by heavy drone production, defensive early game posture, and maximum economy focus.

\item \textbf{Early Pool Aggression}: Features early Zergling pressure with early Spawning Pool (12 pool or earlier) and limited economy buildings. Characterized by sacrificed economy for early pressure and heavy Zergling production.

\item \textbf{+1 Zergling Strategy}: Focuses on melee upgrade with mass Zergling production, with early Evolution Chamber and +1 Melee Attacks. Characterized by timing attacks with upgraded Zerglings and moderate economy into strong timing.

\item \textbf{Nydus-based Strategy}: Utilizes Nydus Network surprise attacks with Lair, Nydus Network, and Nydus Worms. Characterized by map mobility focus, surprise attacks, and specific army compositions.

\item \textbf{Lurker Transition Strategy}: Centers on Lurker-based map control with Lair, Hydralisk Den, and Lurker Den. Characterized by mid-game tech transition, strong map control, and siege capability.

\item \textbf{Early/Mid-early Game All-in Strategy}: Features mass production of basic units with 2-3 bases and relevant production structures without Lair tech. Characterized by heavy unit production, committed aggression, and limited tech investment.
\end{enumerate}

These tactical categories were derived from our corpus analysis as described in  and validated by Grandmaster-level players to ensure they represent the strategic diversity present in high-level StarCraft II gameplay.

\subsection{Model Configurations}
\label{model_config}
We tested several model configurations with different KL divergence weight settings for the action heads to investigate the impact of varying constraints on tactical adaptation. These weights control the strength of the constraint that ties the adapted policy to the original policy, with higher values enforcing stronger adherence to the base policy.

\subsubsection{Configuration A }
This configuration applies uniform constraints across all action heads:
\begin{small}
\begin{verbatim}
head_weights_dict = {
    'action_type': 1.0,
    'delay': 1.0,
    'queued': 1.0,
    'selected_units': 1.0,
    'target_unit': 1.0,
    'target_location': 1.0
}
\end{verbatim}
\end{small}

\subsubsection{Configuration B }
This configuration varies constraints across action heads, with lower constraints on spatial decision-making:
\begin{small}
\begin{verbatim}
head_weights_dict = {
    'action_type': 10.0,
    'delay': 1.0,
    'queued': 10.0,
    'selected_units': 3.0,
    'target_unit': 1.0,
    'target_location': 0.0
}
\end{verbatim}
\end{small}

\subsubsection{Configuration C }
This configuration applies higher uniform constraints across tactical action heads:
\begin{small}
\begin{verbatim}
head_weights_dict = {
    'action_type': 10.0,
    'delay': 1.0,
    'queued': 10.0,
    'selected_units': 10.0,
    'target_unit': 10.0,
    'target_location': 10.0
}
\end{verbatim}
\end{small}

\subsubsection{Configuration D }
This configuration represents maximum constraints across all action heads:
\begin{small}
\begin{verbatim}
head_weights_dict = {
    'action_type': 100.0,
    'delay': 100.0,
    'queued': 100.0,
    'selected_units': 100.0,
    'target_unit': 100.0,
    'target_location': 100.0
}
\end{verbatim}
\end{small}

\section{Discussion}

Our experimental results support three key findings:

\begin{enumerate}
\item \textbf{Tactical Specialization Trade-off}: Higher degrees of tactical adaptation (later epochs) come at the cost of decreased general performance against fixed opponents like the built-in AI. This suggests a natural trade-off between tactical specialization and general capability.

\item \textbf{Constraint Balancing}: The choice of KL divergence weights significantly impacts both the degree of tactical adaptation and the preservation of base capabilities. Configuration C's balanced higher constraints appear to offer a favorable trade-off.

\item \textbf{Adaptation Progression}: The complete dominance of later epoch models over earlier ones in head-to-head competition confirms that meaningful tactical adaptations continue to develop throughout training.
\end{enumerate}

These findings highlight the importance of carefully balancing tactical flexibility and core competency preservation when implementing adapter-based tactical conditioning for complex game AI systems. The optimal configuration depends on the specific application goals—whether prioritizing tactical diversity, competitive performance, or adaptation to specific opponent styles.

\end{document}